\documentclass[letterpaper]{article} 
\usepackage{aaai25}  
\usepackage{times}  
\usepackage{helvet}  
\usepackage{courier}  
\usepackage[hyphens]{url}  
\usepackage{graphicx} 
\usepackage{todonotes}
\usepackage{xcolor}
\usepackage{natbib}  
\usepackage{caption} 
\usepackage{algorithm}
\usepackage{algorithmic}
\usepackage{newfloat}
\usepackage{listings}
\usepackage{amsmath}  
\usepackage{amssymb}  
\usepackage{booktabs} 
\usepackage{multirow} 
\usepackage{subfigure} 
\usepackage{makecell} 
\usepackage{overpic} 
\usepackage{pifont} 
\DeclareCaptionStyle{ruled}{labelfont=normalfont,labelsep=colon,strut=off} 
\floatstyle{ruled}
\newfloat{listing}{tb}{lst}{}
\floatname{listing}{Listing}
\title{
Generalizable Sensor-Based  Activity Recognition via Categorical Concept  \\ Invariant Learning}
\author{
Di~Xiong\textsuperscript{\rm 1}\equalcontrib,
Shuoyuan~Wang\textsuperscript{\rm 2}\equalcontrib,
Lei~Zhang\textsuperscript{\rm 1}\thanks{Corresponding author.},
Wenbo~Huang\textsuperscript{\rm 3},
Chaolei~Han\textsuperscript{\rm 3},
}
\affiliations{
\textsuperscript{\rm 1}Nanjing Normal University, Nanjing 210023, Jiangsu, China\\
\textsuperscript{\rm 2}Southern University of Science and Technology, Shenzhen 518055, Guangdong, China\\
\textsuperscript{\rm 3}Southeast University, Nanjing 211189, Jiangsu, China\\

\{221812013, leizhang\}@njnu.edu.cn, claytonwang0205@gmail.com, \{wenbohuang1002, chaoleihan\}@seu.edu.cn

}

\usepackage{bibentry}

\begin{document}

\maketitle

\begin{abstract}
Human Activity Recognition (HAR) aims to recognize activities by training models on massive sensor data. In real-world deployment, a crucial aspect of HAR that has been largely overlooked is that the test sets may have different distributions from training sets due to inter-subject variability including age, gender, behavioral habits, etc., which leads to poor generalization performance. One promising solution is to learn domain-invariant representations to enable a model to generalize on an unseen distribution. However, most existing methods only consider the feature-invariance of the penultimate layer for domain-invariant learning, which leads to suboptimal results. In this paper, we propose a Categorical Concept Invariant Learning (CCIL) framework for generalizable activity recognition, which introduces a concept matrix to regularize the model in the training stage by simultaneously concentrating on feature-invariance and logit-invariance. Our key idea is that the concept matrix for samples belonging to the same activity category should be similar. Extensive experiments on four public HAR benchmarks demonstrate that our CCIL substantially outperforms the state-of-the-art approaches under cross-person, cross-dataset, cross-position, and one-person-to-another settings.
\end{abstract}
\section{Introduction}
\begin{figure}[t]
	\centering
	\includegraphics[width=0.35\textwidth]{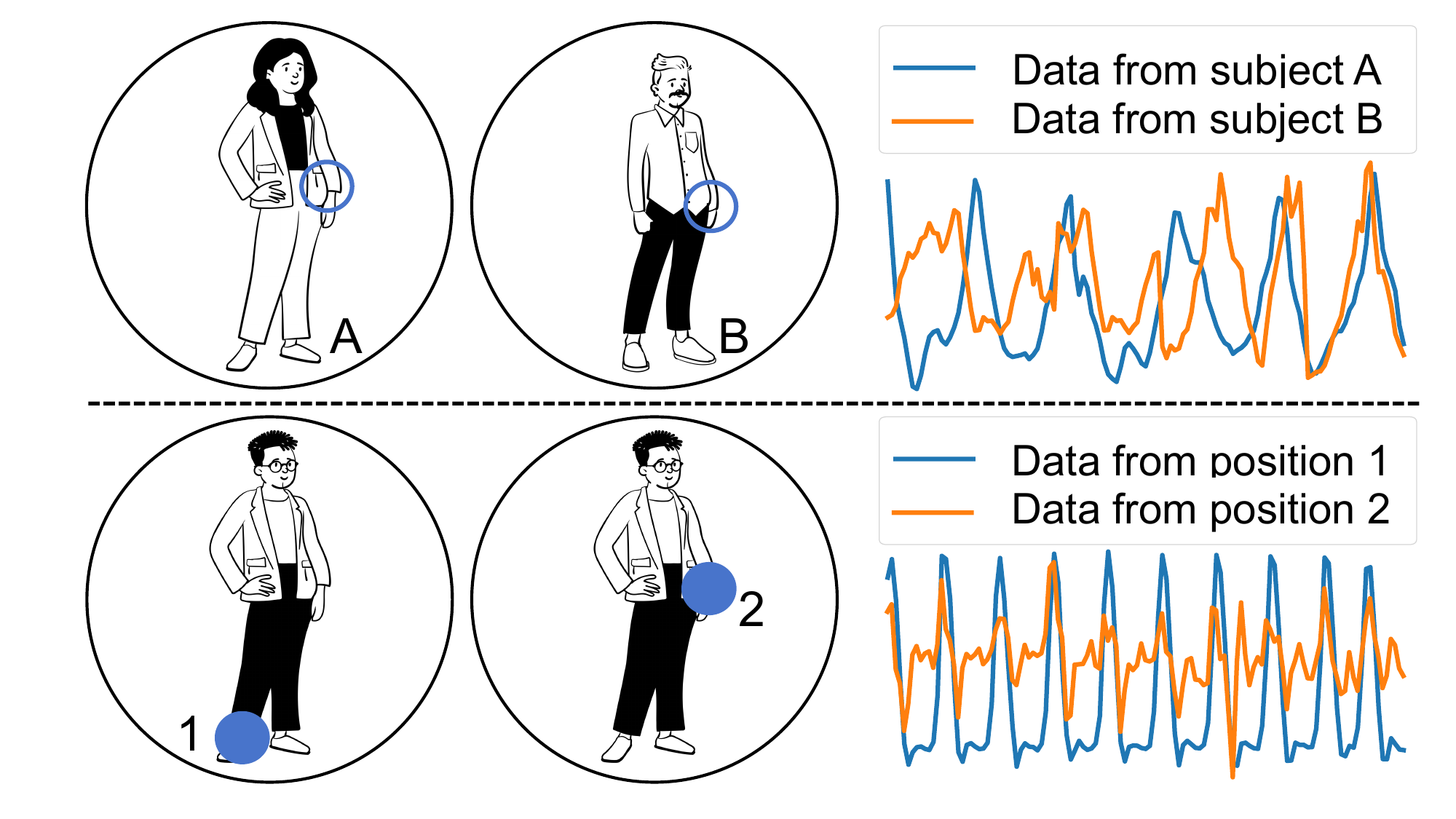} 
	\caption{Domain shift: sensor readings collected from different subjects or different locations of the same subject.}
	\label{fig:top}
\end{figure}

\indent Sensor-based human activity recognition (HAR) aims to train models using massive data collected from wearable sensors such as accelerometers and magnetometers. HAR has wide applications in many areas, including personal fitness, elderly-care, human-machine interaction, sports tracking, etc \cite{huang2022channel}. Despite significant progress, current HAR study still faces critical challenges that prevent practical deployment while rendering the performance suboptimal on never-seen-before data \cite{qian2021latent}. As shown in \figurename~\ref{fig:top}, the distribution of sensor signals is typically influenced by various factors such as age, gender, and deployment locations. For instance, due to inter-subject variability, a model that recognizes the activities of an adult does not generalize well on new unseen data from an elderly person, because they may have different behavioral patterns that make their data distributions highly diverse. Therefore, simply generalizing a model trained on existing data to new unseen data may not work due to such distribution shift problem in sensor signals. \\
\indent In practice, the real-world sensor samples are typically restricted to access during training. Taking elderly fall detection as an example, it is rather unrealistic to aggregate training data from the elderly people based on safety concerns. However, it is feasible to collect training data from young subjects while ensuring enough safe conditions. We have to expect a model trained on the data collected from young subjects to be readily extensible to elderly users with no need of training data collected from them. To mitigate this issue, DG has been a popular technique to reduce the distribution discrepancy between the source and target domains with no need of direct access to never-seen-before target data during training. Though many research efforts have been devoted to computer vision applications, they may be incompatible with time series data. Until now, there has been limited research attention targeted at wearable sensor data including feature disentanglement \cite{qian2021latent}, data augmentation \cite{zhang2018mixup}, gradient operation \cite{huang2020self}, and domain-invariant representation learning \cite{lu2024diversify,du2021adarnn}, which focus on explicitly or implicitly regularizing the models based on the analysis of features. Despite notable achievements in DG, it still remains a major challenge that is far from being solved on sensor data. A recent study \cite{gulrajani2020search} have empirically shown that most current state-of-the-art approaches are even inferior to the baseline empirical risk minimization (ERM) algorithm. These findings clearly highlight the necessity of innovative and effective models that can ensure robustness across domains.\\
\indent Following this cue, as well as the uniqueness of each person’s activity characteristics, this paper takes a different perspective toward obtaining robust outputs through emphasizing the logit-invariance of the classifier weights in a deep learning model, instead of only concentrating on the feature-invariance. As we know, most existing models typically calculate the final output logits through multiplying the classifier weights with the penultimate layer’s feature, where every product term can be viewed as a contribution to the corresponding logit. While organizing all these contributions for logits from all activity classes as a matrix (also named concept matrix) based on input sensor samples, we conjecture that the matrices induced by samples belonging to the same activity class should be similar for a well-generalized model. Based on this intuition, we introduce a new regularization loss term, which aims to enforce similarity between the concept matrix of samples belonging to the same activity category and their corresponding mean value. A dynamic momentum update strategy is used to update the category-wise mean concept matrix during each training iteration. On one hand, different from most existing feature-based regularization \cite{cha2022domain}, our approach also takes into full consideration the effect of the classifier weights, avoiding biased estimation for feature importance. On the other hand, to overcome the drawback of logit-based regularization that only has a coarse value, our approach provides a fine-grained characterization of cross-domain activity recognition by considering the varying influence of every contribution to final classification results. In summary, the main contributions of this paper are three-fold:
\begin{itemize}
    \item \textbf{New perspective:} In this paper, we propose a Categorical Concept Invariant Learning framework named CCIL, which is mainly built on the category-wise mean concept matrix. We provide new insights from both logit-invariance and feature-invariance perspectives to explain the rationale behind our generalizable activity recognition algorithm.
    \item \textbf{Simple algorithm:} A new regularization term is introduced to enforce similarity between the concept matrix of samples from the same activity class and their mean value, while a dynamic momentum update strategy is used to update the matrix during each training iteration. Our approach is simple, which adds only a few lines of code upon the standard ERM baseline. 
    \item \textbf{Superior performance:} Comprehensive experiments on four public sensor-based HAR datasets demonstrate that our proposed CCIL consistently beat the state-of-the-art baselines while evaluated under the rigorous cross-domain settings. These results highlight the effectiveness and universality of our concept matrix invariance regularization, despite its simplicity. 
\end{itemize}
\section{Related Work}
\subsection{Human Activity Recognition}
 \indent Human activity recognition (HAR) mainly attempts to recognize activities of daily living that are performed by different persons. Based on data type, it can be roughly grouped into two categories: vision-based HAR and sensor-based HAR \cite{2020Sensor}: The former collects activity data by cameras or other optical devices, which would often encounter severe privacy leaking problems \cite{sun2022human}. For example, sensitive personal data like facial information will be accidentally released on cameras. Moreover, cameras may not work in HAR when a person is beyond their coverage range \cite{kong2022human}; The latter collects activity data through ambient sensors deployed in smart environment or wearable sensors attached to different body parts \cite{2021A}. Due to small size and low price, there has been the wide popularity of inertial sensors embedded in wearable devices like smart phones and watches, that makes them convenient and practical to record activity data for offering smart user services \cite{chen2021deep}. In contrast to video-based action recognition, there is relative limited research attention on HAR using wearable sensor data. Thus, this paper mainly concentrates on wearable sensor-based HAR problem. To resolve sensor-based HAR, deep learning models have recently been widely applied to automatically extract features from raw sensor signals for activity recognition \cite{qian2021latent,hammerla2016deep,2019A,wang2024optimization}. Despite remarkable progress, these activity recognition models are typically trained based on the assumption that the training and testing data have independently identical distributions, ignoring the fact the sensor data collected from different persons may follow different distributions due to their unique characteristics in body shapes, behavior patterns, or other biological factors.
 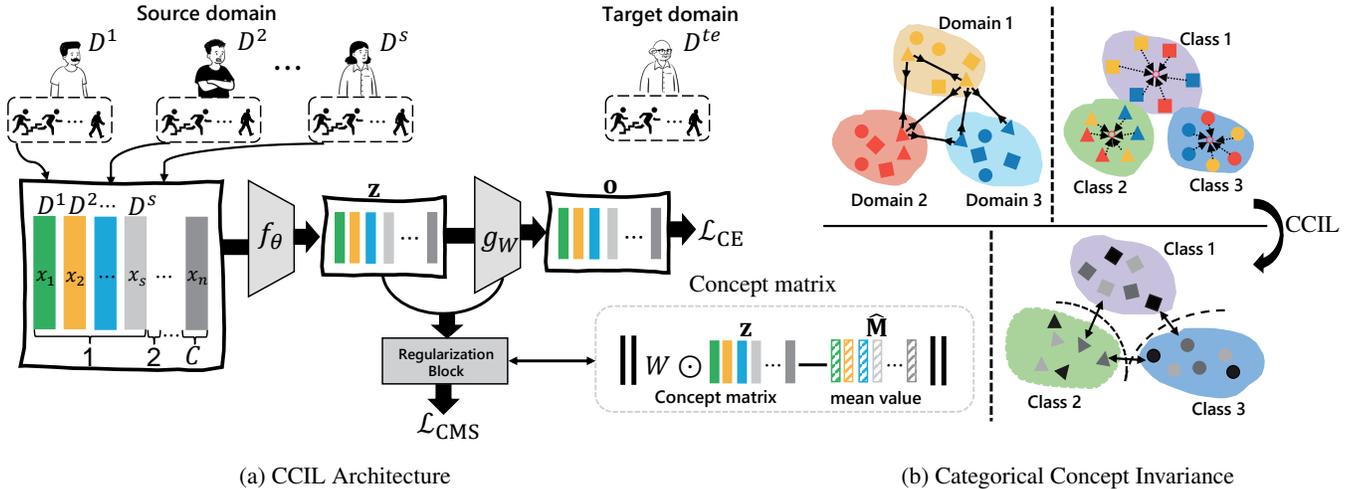
\begin{figure*}[ht]
	\centering
	\begin{overpic}[width=0.99\textwidth]{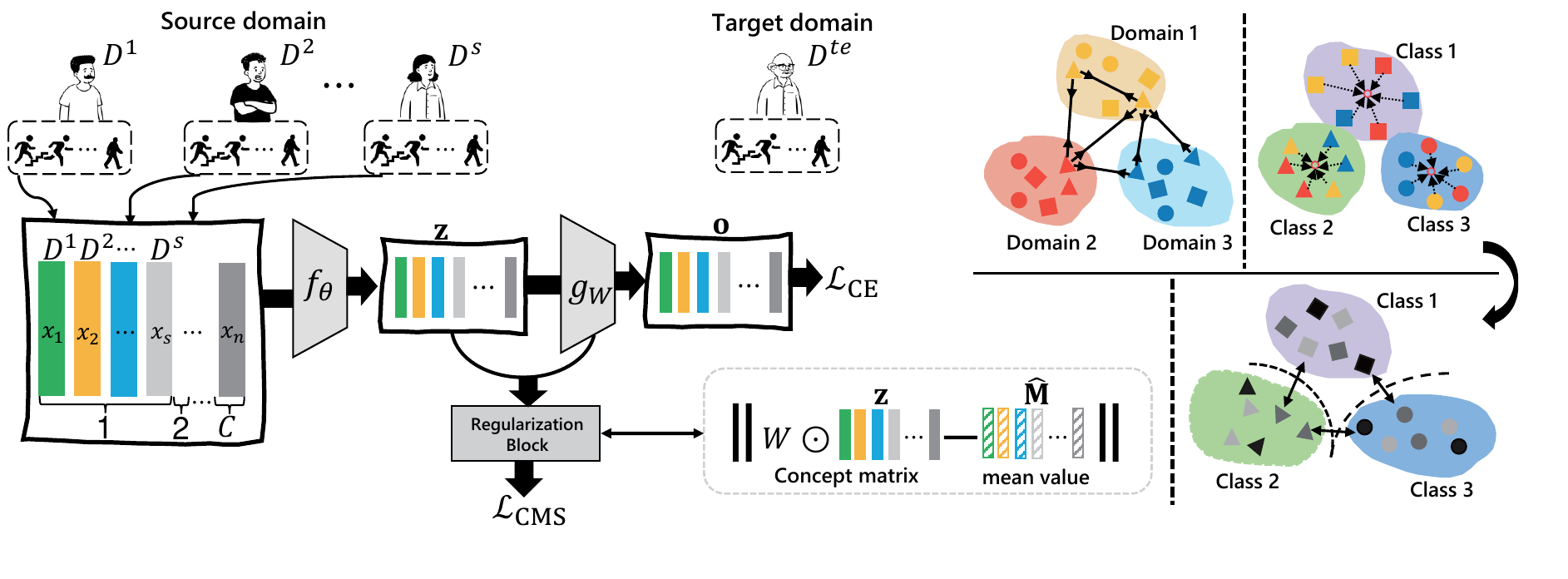}
        \put(97, 19){\footnotesize {\textcolor{black}{CCIL}}}
        \put(52, 14.5){\footnotesize {\textcolor{black}{Concept matrix}}}
        \put(18, 0){\footnotesize {\textcolor{black}{(a) CCIL Architecture}}}
        \put(68, 0){\footnotesize {\textcolor{black}{(b) Categorical Concept Invariance}}}
	\end{overpic}
	\caption{(a) An overview of our CCIL framework based on the concept matrix. (b) CCIL learns domain-invariant representation by mapping the latent representation of the same-class samples close together}
	\label{fig1}
\end{figure*}
\subsection{Domain-Invariant Learning for Generalization}
\indent To address distribution-shift problem, domain adaptation (DA) has offered a popular solution to bridge domain gaps \cite{kouw2019review,wang2018stratified,yu2023semi}. However, DA has a notable limitation in that it more or less requires to access target domain data during training, which renders DA infeasible in many real-world HAR situations \cite{qian2021latent}. Moreover, while there are multiple target domains, the model has to be re-trained on every target domain \cite{lu2021cross}, which is time-consuming and inefficient. DG has been recently proposed to handle such challenging situations \cite{wang2018stratified}. It aims to learn a robust and generalizable model from one or more different but related source domains, that can perform well on the never-seen-before target domains. In computer vision community, existing DG-related literatures may be coarsely categorized into three research streams \cite{zhou2022domain,wang2022generalizing}: Learning strategy \cite{huang2020self,sagawadistributionally}, Data manipulation \cite{zhang2018mixup,zhoudomain}, Representation learning \cite{parascandololearning,du2021adarnn,qian2021latent,lu2024diversify,chen2023domain,sun2016deep}. Though the past five years have witnessed its remarkable success in the computer vision community, there has been very limited research attention for human activity recognition using wearable sensor data. To the best of our knowledge, the latest work to solve such DG problem for HAR is DIVERSITY \cite{lu2024diversify}, which leverages the latent distributions to minimize the distribution divergence for time series out-of-distribution detection and generalization. There now exists very few DG-related works that explore the effect of classifier weights on series time data for cross-domain activity recognition. Different from prior arts, our work is the first to simultaneously concentrate on the feature and logit invariance regularizations, highlighting the effectiveness of the concept matrix.

\section{Methodology}
\subsection{Problem Formulation}
Given a set of source domains utilized as the training dataset $\mathcal{D}^{tr} =\{\mathcal{D}^1,\mathcal{D}^2,... ,\mathcal{D}^S\}$, for the $i$-th source domain $\mathcal{D}^i$ we use $P^i\left(\boldsymbol{x},y\right)$ on $\mathcal{X}\times\mathcal{Y}$ to represent the joint distribution, where $\boldsymbol{x}\in\mathcal{X}$ denotes the sensor input obtained by sliding window \footnote{Due to its implementational simplicity, a popuar sliding window strategy is used to divide continous time series data into fixed-size windows as sensor inputs here \cite{ordonez2016deep}.}, $y\in \mathcal{Y}=\{1,... ,C\}$ denotes the label space with total $C$ activity categories. We perform training from the source domain $\mathcal{D}^{tr}$ dataset to obtain the cross-domain activity recognition model $h:\mathcal{X} \rightarrow \mathcal{Y}$, which can generalize well on the never-seen-before target domain $\mathcal{D}^{te}$ dataset. It is important to note that the target domain $\mathcal{D}^{te}$ can only be accessed at inference. Although the source and target domains have the same input and output spaces, the source and target domains have different distributions $P^{te}$, i.e., ${{P}^{i}}(\boldsymbol{x},y)\ne {{P}^{j}}(\boldsymbol{x},y),\forall i,j\in \{1,2,...,S,te\}$. In a word, we hope that the model $h$ trained on the source domain $D^{tr}$ can minimize the average prediction error $\epsilon_t$ on the target domain $D^{te}$:
\begin{equation}
	\epsilon_t = \mathbb{E}_{(\boldsymbol{x}, y) \sim {P}^{te}(\boldsymbol{x}, y)}\mathcal{L} (h(\mathbf{x}), y).
	\label{eq:1}
\end{equation}

\subsection{Motivation Based on Feature v.s. Logit}
\indent In this section, we answer the potential motivation behind the proposed CCIL and provide in-depth insights into our algorithm design. To learn domain-invariant representations for sensor-based activity recognition, a well-generalized model should be stable under cross-domain scenarios. Most existing DG-based works concentrate on implicitly or explicitly regularizing the model based on the notation of feature-invariance \cite{lu2024diversify,du2021adarnn,cha2022domain}. However, such feature-invariance leaning strategy still poses a serious limitation. To be specific, while only focusing on the feature-invariance, it fails to take into full consideration the classifier weights, which are in charge of determining the importance of different feature elements, thereby resulting in a biased estimate for feature importance. For example, while a feature element has a big value, it might correspond to a small value in the classifier, leading to a lower effect on final activity classification results. Solely considering the feature-invariance would be biased or misleading, thus undermining the generalization ability of the model. Therefore, the influence of the classifier weights should be fully considered so as to avoid such biased estimation of feature importance.\\
\indent To mitigate this issue in cross-domain situation, instead of only concentrating on the feature-invariance, the logit may implicitly take into account the effect of classifier weights to a certain extent. However, the logit is only able to provide a coarse value, that lacks a fine-grained perspective to interpret the rationale behind generalizable cross-domain activity recognition process. As a consequence, focusing only on logit-invariance may lead to ineffectiveness in generating robust feature representations, which will be verified in later visualizing analysis. Our concept matrix aims to overcome the two drawbacks by simultaneously concentrating on both feature-invariance or logit-invariance, which learn domain-invariant representations from a more fine-grained perspective while taking into account the influence of classifier weights. A new regularization loss term is introduced based on the concept matrix to capture both feature-invariance and logit-invariance representations for generalizable cross-domain activity recognition.
\subsection{Framework Overview}
\indent This section presents a comprehensive description of our newly proposed DG approach. As illustrated in \figurename~\ref{fig1}, we propose Categorical Concept Invariant Learning abbreviated as CCIL to learn both feature-invariance and logit-invariance for generalizable cross-domain activity recognition. CCIL takes inputs from multiple different but related source domains for model training, while the target domain data is only utilized for model test. Subsequently, after going through a common feature extractor, the output features are multiplied with the classifier weights, which can then be used to form the concept matrix. While the same activity performed by different persons (domains) tend to have similar activity semantics, we may leverage the invariance across domains to regularize the model to facilitate domain generalization. To ensure robust results, the concept matrix of samples belonging to the same activity class should align with their corresponding mean value, which is very reasonable such the causal factors for invariance-learning are usually stable patterns to persist across domains \cite{lu2024diversify,chen2023domain}. On this basis, we introduce a new regularization loss term that allows the model to explore more invariance.
\subsubsection{Concept Matrix}
In most existing HAR models, the backbone architecture $h$ is typically comprised of a feature extractor and an activity classifier. The feature $\mathbf{z} \in \mathbb{R}^D$ can be produced through a feature extractor $f$ (i.e., \ $\mathbf{z}={f}\left(x\right)$) parameterized by $\theta$, which contains two convolution layers and one pooling layer \cite{lu2024diversify}. Assuming that there are total $C$ activity classes in $\mathcal{Y}$ after applying the classifier $g$ comprised of one fully-connected layer on $\mathbf{z}$, we can obtain the final logits $\mathbf{o}=\mathbf{W}^\top\ \mathbf{z}\in\mathbb{R}^C$ (i.e., $\mathbf{o}=g\left(\mathbf{z}\right)$), where $\mathbf{W}\in\mathbb{R}^{D\times C}$ is the weights of the classifier $g$. For simplicity, we omit the bias in the classifier. On this basis, the concept matrix can be constructed based on the output feature $\mathbf{z}$ and classifier weights $\mathbf{W}$. In practice, every logit value is equivalent to the summation of element-wise multiplications between the feature elements and the corresponding weights in the classifier. Without loss of generality, $o_c$ (i.e., the $c$-th dimension of $\mathbf{o}$) can be formulated as follows: 
\begin{equation}
	{o_c=\mathbf{W}}^\top_{\{,c\}}=\sum_{j=1}^{D}W_{{\{j,c\}}}{z_j},
	\label{eq:2}
\end{equation}
where the logit value on the $c$-th activity class is a simple addition of all $W_{\left\{j,c\right\}}z_j$. Intuitively, it can be seen as an element-wise contribution to $o_c$. Therefore, we are able to aggregate all $W_{\left\{j,c\right\}}z_j$ to form the concept matrix, that may be mathematically denoted as follows:
\begin{equation}
	\mathbf{M}=\left[\begin{matrix}W_{\{1,1\}}z_1&W_{\{1,2\}}z_1&\ldots&W_{\{1,C\}}z_1\\W_{\{2,1\}}z_2&W_{\{2,2\}}z_2&\ldots&W_{\{2,C\}}z_2\\\vdots&\vdots&\ddots&\vdots\\W_{\{D,1\}}z_D&W_{\{D,2\}}z_D&\ldots&W_{\{D,C\}}z_D\\\end{matrix}\right].
	\label{eq:3}
\end{equation}
Since the Softmax function is implemented on all the logits to produce the final posterior probability. It is important to note that the concept matrix $\mathbf{M}$ should be constructed from all classes. That is to say, the final posterior probability will be affected by the logits from all activity classes.
\subsubsection{Categorical Concept Invariant Learning}
Our key idea is that the concept matrix for activity samples of the same category should align with their corresponding mean value, implying that the concept matrix for the same activity category should be similar regardless of domains. To achieve this goal, we introduce a regularization term based on the concept matrix similarity (abbreviated as CMS) in training phase. Such regularization term may be formulated as:
\begin{equation}
	\mathcal{L}_\mathrm{CMS}=\frac{1}{N_{b}}\sum_{c}\sum_{\{i|y_{i}=c\}}\|\mathbf{M}_{i}-\hat{\mathbf{M}}_{c}\|^{2},
	\label{eq:4}
\end{equation}
where $N_b$ is the number of samples in one mini-batch, $\mathbf{M}_i$ denotes the concept matrix of the $i$-th sample, $\hat{\mathbf{M}}_{c}$ denotes the mean matrix of the concept matrix corresponding to the $c$-th class, and $\Vert \cdot \Vert$ is $l_2$ norm.
The $\hat{\mathbf{M}}_{c}$ in the above equation requires averaging the activity samples in all domains, which is impractical and computationally expensive. Since it is unrealistic to directly calculate the concept matrix of all sensor samples, we perform a dynamic momentum update to adapt the concept matrix during each training iteration, which can greatly ease the computational burden. To be specific, inspired by previous work \cite{he2020momentum,chen2023domain} we can utilize momentum updating for $\hat{\mathbf{M}}_{c}$ online:
\begin{equation}
	\hat{\mathbf{M}}_{c}^{t}=(1-\lambda)\times\hat{\mathbf{M}}_{c}^{t-1}+\lambda \times\frac{1}{|y_{i}=c|}\sum_{\{i|y_{i}=c\}}\mathbf{M}_{i},
	\label{eq:5}
\end{equation}
where $\lambda$ is the positive momentum value, $t$ is the iteration index, $|y_{i}=c|$ denotes the sample corresponding to the $c$-th class of activity identification, and $\hat{\mathbf{M}}_{c}$ is initialized from the first iteration to compute the processed concept matrix.
\subsubsection{Learning Objective}
The overall learning objective can be written as follows:
\begin{equation}
	\mathcal{L} = \mathcal{L}_\mathrm{CE} + \alpha \mathcal{L}_\mathrm{CMS}, 
	\label{eq:6}
\end{equation}
where $\mathcal{L}_\mathrm{CE}$ denotes the standard cross-entropy loss, $\mathcal{L}_\mathrm{CMS}$ is the loss of CCIL, and $\alpha$ is a positive weight coefficient. As can be seen in Eq. \ref{eq:6}, our CCIL is very simple, that only requires adding only a few lines of code upon the vanialla ERM training pipeline.
\begin{table*}[ht]
	\small
	\centering
	\begin{tabular}{l|ccccc|ccccc|ccccc}
		\Xhline{1pt}
		\multirow{2}{*}{Method} & \multicolumn{5}{c|}{Target (DSADS)} & \multicolumn{5}{c|}{Target (USC-HAD)} & \multicolumn{4}{c}{Target (PAMAP2)} \\ \cline{2-16} 
		& \multicolumn{1}{c|}{0} & \multicolumn{1}{c|}{1} & \multicolumn{1}{c|}{2} & \multicolumn{1}{c|}{3} & AVG & \multicolumn{1}{c|}{0} & \multicolumn{1}{c|}{1} & \multicolumn{1}{c|}{2} & \multicolumn{1}{c|}{3} & AVG & \multicolumn{1}{c|}{0} & \multicolumn{1}{c|}{1} & \multicolumn{1}{c|}{2} & \multicolumn{1}{c|}{3} & AVG \\ \hline
		ERM & \multicolumn{1}{c|}{83.1} & \multicolumn{1}{c|}{79.3} & \multicolumn{1}{c|}{87.8} & \multicolumn{1}{c|}{71.0} & 80.3 & \multicolumn{1}{c|}{81.0} & \multicolumn{1}{c|}{57.7} & \multicolumn{1}{c|}{74.0} & \multicolumn{1}{c|}{65.9} & 69.7 & \multicolumn{1}{c|}{90.0} & \multicolumn{1}{c|}{78.1} & \multicolumn{1}{c|}{55.8} & \multicolumn{1}{c|}{84.4} & 77.1  \\
		DANN & \multicolumn{1}{c|}{89.1} & \multicolumn{1}{c|}{84.2} & \multicolumn{1}{c|}{85.9} & \multicolumn{1}{c|}{83.4} & 85.6 & \multicolumn{1}{c|}{81.2} & \multicolumn{1}{c|}{57.9} & \multicolumn{1}{c|}{76.7} & \multicolumn{1}{c|}{70.7} & 71.6 & \multicolumn{1}{c|}{82.2} & \multicolumn{1}{c|}{78.1} & \multicolumn{1}{c|}{55.8} & \multicolumn{1}{c|}{87.3} & 75.7  \\
		CORAL & \multicolumn{1}{c|}{91.0} & \multicolumn{1}{c|}{85.8} & \multicolumn{1}{c|}{86.6} & \multicolumn{1}{c|}{78.2} & 85.4 & \multicolumn{1}{c|}{78.8} & \multicolumn{1}{c|}{58.9} & \multicolumn{1}{c|}{75.0} & \multicolumn{1}{c|}{53.7} & 66.6 & \multicolumn{1}{c|}{86.2} & \multicolumn{1}{c|}{77.8} & \multicolumn{1}{c|}{49.0} & \multicolumn{1}{c|}{\underline{87.8}} & 75.2  \\
		Mixup & \multicolumn{1}{c|}{89.6} & \multicolumn{1}{c|}{82.2} & \multicolumn{1}{c|}{89.2} & \multicolumn{1}{c|}{\underline{86.9}} & 87.0 & \multicolumn{1}{c|}{80.0} & \multicolumn{1}{c|}{\underline{64.1}} & \multicolumn{1}{c|}{74.3} & \multicolumn{1}{c|}{61.3} & 69.9 & \multicolumn{1}{c|}{89.4} & \multicolumn{1}{c|}{80.3} & \multicolumn{1}{c|}{58.4} & \multicolumn{1}{c|}{87.7} & 79.0  \\
		GroupDRO & \multicolumn{1}{c|}{\underline{91.7}} & \multicolumn{1}{c|}{85.9} & \multicolumn{1}{c|}{87.6} & \multicolumn{1}{c|}{78.3} & 85.9 & \multicolumn{1}{c|}{80.1} & \multicolumn{1}{c|}{55.5} & \multicolumn{1}{c|}{74.7} & \multicolumn{1}{c|}{60.0} & 67.6 & \multicolumn{1}{c|}{85.2} & \multicolumn{1}{c|}{77.7} & \multicolumn{1}{c|}{56.2} & \multicolumn{1}{c|}{85.0} & 76.0 \\
		RSC & \multicolumn{1}{c|}{84.9} & \multicolumn{1}{c|}{82.3} & \multicolumn{1}{c|}{86.7} & \multicolumn{1}{c|}{77.7} & 82.9 & \multicolumn{1}{c|}{81.9} & \multicolumn{1}{c|}{57.9} & \multicolumn{1}{c|}{73.4} & \multicolumn{1}{c|}{65.1} & 69.6 & \multicolumn{1}{c|}{87.1} & \multicolumn{1}{c|}{76.9} & \multicolumn{1}{c|}{{60.3}} & \multicolumn{1}{c|}{\underline{87.8}} & 78.0\\
		ANDMask & \multicolumn{1}{c|}{85.0} & \multicolumn{1}{c|}{75.8} & \multicolumn{1}{c|}{87.0} & \multicolumn{1}{c|}{77.6} & 81.4 & \multicolumn{1}{c|}{79.9} & \multicolumn{1}{c|}{55.3} & \multicolumn{1}{c|}{74.5} & \multicolumn{1}{c|}{65.0} & 68.7 & \multicolumn{1}{c|}{86.7} & \multicolumn{1}{c|}{76.4} & \multicolumn{1}{c|}{43.6} & \multicolumn{1}{c|}{85.6} & 73.1 \\
		GILE & \multicolumn{1}{c|}{81.0} & \multicolumn{1}{c|}{75.0} & \multicolumn{1}{c|}{77.0} & \multicolumn{1}{c|}{66.0} & 74.7 & \multicolumn{1}{c|}{78.0} & \multicolumn{1}{c|}{62.0} & \multicolumn{1}{c|}{77.0} & \multicolumn{1}{c|}{63.0} & 70.0 & \multicolumn{1}{c|}{83.0} & \multicolumn{1}{c|}{68.0} & \multicolumn{1}{c|}{42.0} & \multicolumn{1}{c|}{76.0} & 67.5 \\
		AdaRNN & \multicolumn{1}{c|}{80.9} & \multicolumn{1}{c|}{75.5} & \multicolumn{1}{c|}{\underline{90.2}} & \multicolumn{1}{c|}{75.5} & 80.5 & \multicolumn{1}{c|}{78.6} & \multicolumn{1}{c|}{55.3} & \multicolumn{1}{c|}{66.9} & \multicolumn{1}{c|}{\underline{73.7}} & 68.6 & \multicolumn{1}{c|}{81.6} & \multicolumn{1}{c|}{71.8} & \multicolumn{1}{c|}{45.4} & \multicolumn{1}{c|}{82.7} & 70.4 \\
		DIVERSIFY & \multicolumn{1}{c|}{90.4} & \multicolumn{1}{c|}{\underline{86.5}} & \multicolumn{1}{c|}{90.0} & \multicolumn{1}{c|}{86.1} & \underline{88.2} & \multicolumn{1}{c|}{\underline{82.6}} & \multicolumn{1}{c|}{63.5} & \multicolumn{1}{c|}{\underline{78.7}} & \multicolumn{1}{c|}{71.3} & \underline{74.0} & \multicolumn{1}{c|}{\underline{91.0}} & \multicolumn{1}{c|}{\underline{84.3}} & \multicolumn{1}{c|}{\underline{60.5}} & \multicolumn{1}{c|}{87.7} & \underline{80.8}  \\ \hline
		Ours & \multicolumn{1}{c|}{\textbf{94.7}} & \multicolumn{1}{c|}{\textbf{88.2}} & \multicolumn{1}{c|}{\textbf{92.5}} & \multicolumn{1}{c|}{\textbf{87.5}} & \textbf{90.7} & \multicolumn{1}{c|}{\textbf{85.2}} & \multicolumn{1}{c|}{\textbf{66.5}} & \multicolumn{1}{c|}{\textbf{79.3}} & \multicolumn{1}{c|}{\textbf{77.0}} & \textbf{77.0} & \multicolumn{1}{c|}{\textbf{93.8}} & \multicolumn{1}{c|}{\textbf{87.2}} & \multicolumn{1}{c|}{\textbf{63.8}} & \multicolumn{1}{c|}{\textbf{93.2}} & \textbf{84.5} \\ 
		\Xhline{1pt}
	\end{tabular}
	\caption{Accuracy on cross-person generalization. We use 0,1,2,3 to denotes the unseen test set. \textbf{Bold} means the best while \underline{underline} means the second-best.}
	\label{tab:1}
\end{table*}
\begin{table}[ht]
	\small  
	\centering
	\begin{tabular}{l|cccccc}
		\Xhline{1pt}
		\multirow{2}{*}{Method} & \multicolumn{6}{c}{Target (DSADS)} \\ \cline{2-7} 
		& \multicolumn{1}{c|}{0} & \multicolumn{1}{c|}{1} & \multicolumn{1}{c|}{2} & \multicolumn{1}{c|}{3} & \multicolumn{1}{c|}{4} & AVG \\ \hline
		ERM & \multicolumn{1}{c|}{41.5} & \multicolumn{1}{c|}{26.7} & \multicolumn{1}{c|}{35.8} & \multicolumn{1}{c|}{21.4} & \multicolumn{1}{c|}{27.3} & 30.6 \\
		DANN & \multicolumn{1}{c|}{45.4} & \multicolumn{1}{c|}{25.3} & \multicolumn{1}{c|}{38.1} & \multicolumn{1}{c|}{28.9} & \multicolumn{1}{c|}{25.1} & 32.6 \\
		CORAL & \multicolumn{1}{c|}{33.2} & \multicolumn{1}{c|}{25.2} & \multicolumn{1}{c|}{25.8} & \multicolumn{1}{c|}{22.3} & \multicolumn{1}{c|}{20.6} & 25.4 \\
		Mixup & \multicolumn{1}{c|}{\underline{48.8}} & \multicolumn{1}{c|}{\underline{34.2}} & \multicolumn{1}{c|}{37.5} & \multicolumn{1}{c|}{29.5} & \multicolumn{1}{c|}{29.9} & 36.0 \\
		GroupDRO & \multicolumn{1}{c|}{27.1} & \multicolumn{1}{c|}{26.7} & \multicolumn{1}{c|}{24.3} & \multicolumn{1}{c|}{18.4} & \multicolumn{1}{c|}{24.8} & 24.3 \\
		RSC & \multicolumn{1}{c|}{46.4} & \multicolumn{1}{c|}{27.4} & \multicolumn{1}{c|}{35.9} & \multicolumn{1}{c|}{27.0} & \multicolumn{1}{c|}{29.8} & 33.3 \\
		ANDMask & \multicolumn{1}{c|}{47.5} & \multicolumn{1}{c|}{31.1} & \multicolumn{1}{c|}{39.2} & \multicolumn{1}{c|}{30.2} & \multicolumn{1}{c|}{29.9} & 35.6 \\
		DIVERSIFY & \multicolumn{1}{c|}{47.7} & \multicolumn{1}{c|}{32.9} & \multicolumn{1}{c|}{\textbf{44.5}} & \multicolumn{1}{c|}{\textbf{31.6}} & \multicolumn{1}{c|}{\underline{30.4}} & \underline{37.4} \\ \hline
		Ours & \multicolumn{1}{c|}{\textbf{49.6}} & \multicolumn{1}{c|}{\textbf{35.6}} & \multicolumn{1}{c|}{\underline{44.2}} & \multicolumn{1}{c|}{\underline{31.4}} & \multicolumn{1}{c|}{\textbf{32.6}} & \textbf{38.7} \\ 
		\Xhline{1pt}
	\end{tabular}
	\caption{Accuracy on cross-position generalization. We use 0,1,2,3,4 to denotes the unseen test set. \textbf{Bold} means the best while \underline{underline} means the second-best.}
	\label{tab:2}
\end{table}
\section{Experiments}
\subsection{Experimental Setup}
\subsubsection{Dataset and Model Architecture}
The widely-employed sliding window strategy is first used to segment time series data, while maintaining the same window length and overlap rate as in previous works \cite{ordonez2016deep,anguita2013public,wang2019deep}. We directly follow the model architecture in \cite{lu2024diversify} to conduct the experiments. The backbone architecture consists of two modules: the feature extractor and activity classifier. The feature extractor includes two convolutional layers followed by max-pooling operation for feature extraction, while the classifier contains a fully connected layer for final predictions. We evaluate our method on four public sensor-based HAR benchmark: DSADS \cite{altun2010comparative}, PAMAP2 \cite{reiss2012introducing}, USC-HAD \cite{zhang2012usc} and UCI-HAR \cite{anguita2013public}.

\subsubsection{Cross-Domain Settings} Cross-domain Settings are divided into the following four categories. \emph{Cross-person} setting \footnote{Since the baselines for UCI-HAR are already good enough, we do not run cross-person experiments on it.}. In the DSADS dataset, there are a total of 8 subjects. We divide the 8 subjects into 4 domains, each of which contains two subjects. We use the sliding window technique with a window size of 125 and an overlap rate of 50\%. The final processed sample size is (45, 1, 125), where 45 represents sensors from 5 positions, with each position having 3 different sensors, and each sensor being 3-axis. In the USC-HAD dataset, there are a total of 14 subjects. We roughly divide them into four domains, where three of four domains with each containing four subjects are used as source domain, while the rest domain containing two subjects is utilized as target domain. We use the sliding window technique with a window size of 200 and an overlap rate of 50\%. The final processed sample size is (6, 1, 200), where 6 represents sensors from one position, with this position having 2 different sensors, and each sensor being 3-axis. In the PAMAP2 dataset, there are a total of 9 subjects with subject IDs 0–8. We divide them into four domains: domains: (2, 3, 8), (1, 5), (0, 7), (4, 6). We use the sliding window technique with a window size of 200 and an overlap rate of 50\%. The final processed sample size is (27, 1, 200), where 27 represents sensors from 3 positions, with each position having 3 different sensors, and each sensor being 3-axis; \emph{Cross-position} setting. We utilize the DSADS dataset for cross-position experiments, dividing it into five domains based on position. The sliding window size and overlap rate are consistent with those in the cross-person setting. The final processed sample size is (9, 1, 125), where 9 represents three sensors from a single position, with each sensor capturing three-axis data; \emph{Cross-dataset} setting. We merge four datasets, which are then roughly divided  into four domains. We select six common activities across all four datasets which come from two sensors at similar or identical positions in each dataset. The high-frequency datasets such as USC-HAD are downsampled to ensure consistent sensor sampling frequencies for alignment. The sliding window size and overlap rate are the same as those in the cross-person setting, resulting in a final processed sample size of (6, 1, 50); \emph{One-person-to-another} setting. We utilize the DSADS, USC-HAD, and PAMAP2 datasets, selecting four pairs of subjects to generalize from one subject to another. Specifically, the pairs are (0, 1), (2, 3), (4, 5), and (6, 7), where we generalize from the second subject in each pair to the first. The sliding window size, overlap rate, and final processed sample size are consistent with those in the cross-person setting.

\subsubsection{Comparative Methods}
We compare our approach with three recent methods: GILE \cite{qian2021latent}, AdaRNN \cite{du2021adarnn}, and DIVERSIFY \cite{lu2024diversify}. We will also compare it with seven commonly used domain generalization (DG) methods: EMR \cite{vapnik1991principles}, DANN \cite{ganin2016domain}, CORAL \cite{sun2016deep}, Mixup \cite{zhang2018mixup}, GroupDRO \cite{sagawadistributionally}, RSC \cite{huang2020self}, and ANDMask \cite{parascandololearning}. For a fair comparison, all methods, except GILE and AdaRNN, use the same network architecture.
\begin{table}[t]
	\small
	\centering
	\begin{tabular}{l|ccccc}
		\Xhline{1pt}
		\multirow{2}{*}{Method} & \multicolumn{5}{c}{Target} \\ \cline{2-6} 
		& \multicolumn{1}{c|}{0} & \multicolumn{1}{c|}{1} & \multicolumn{1}{c|}{2} & \multicolumn{1}{c|}{3} & AVG \\ \hline
		ERM & \multicolumn{1}{c|}{26.4} & \multicolumn{1}{c|}{29.6} & \multicolumn{1}{c|}{44.4} & \multicolumn{1}{c|}{32.9} & 33.3 \\
		DANN & \multicolumn{1}{c|}{29.7} & \multicolumn{1}{c|}{45.3} & \multicolumn{1}{c|}{46.1} & \multicolumn{1}{c|}{43.8} & 41.2 \\
		CORAL & \multicolumn{1}{c|}{39.5} & \multicolumn{1}{c|}{41.8} & \multicolumn{1}{c|}{39.1} & \multicolumn{1}{c|}{36.6} & 39.2 \\
		Mixup & \multicolumn{1}{c|}{37.3} & \multicolumn{1}{c|}{47.4} & \multicolumn{1}{c|}{40.2} & \multicolumn{1}{c|}{23.1} & 37.0 \\
		GroupDRO & \multicolumn{1}{c|}{\underline{51.4}} & \multicolumn{1}{c|}{36.7} & \multicolumn{1}{c|}{33.2} & \multicolumn{1}{c|}{33.8} & 38.8 \\
		RSC & \multicolumn{1}{c|}{33.1} & \multicolumn{1}{c|}{39.7} & \multicolumn{1}{c|}{45.3} & \multicolumn{1}{c|}{45.9} & 41.0 \\
		ANDMask & \multicolumn{1}{c|}{41.7} & \multicolumn{1}{c|}{33.8} & \multicolumn{1}{c|}{43.2} & \multicolumn{1}{c|}{40.2} & 39.7 \\
		DIVERSIFY & \multicolumn{1}{c|}{48.7} & \multicolumn{1}{c|}{\underline{46.9}} & \multicolumn{1}{c|}{\underline{49.0}} & \multicolumn{1}{c|}{\textbf{59.9}} & \underline{51.1} \\ \hline
		Ours & \multicolumn{1}{c|}{\textbf{52.1}} & \multicolumn{1}{c|}{\textbf{48.5}} & \multicolumn{1}{c|}{\textbf{50.3}} & \multicolumn{1}{c|}{\underline{59.6}} & \textbf{52.6} \\ 
		\Xhline{1pt}
	\end{tabular}
	\caption{Accuracy on cross-dataset generalization. We use 0,1,2,3 to denotes the unseen test set. 0 represents DSADS, 1 represents USC-HAD, 2 represents UCI-HAR, and 3 represents PAMAP2. \textbf{Bold} means the best while \underline{underline} means the second-best.}
	\label{tab:3}
\end{table}
\subsubsection{Implementation Details}
The maximum training period was set to 150 epochs and an Adam optimizer with a weight decay of $5\times{10^{-4}}$ was used. All methods utilized a learning rate of ${10^{-2}}$ or ${10^{-3}}$. In all experiments, the batch size was set to 32. Some DG methods require domain labels to be known during training, whereas ours do not, making our approach both more challenging and more practical. For methods that require domain labels, we assigned domain labels in batches. Following the generalization setup of HAR in DIVERSIFY \cite{lu2024diversify}, we employed a source-domain validation strategy. The source domain data was split into training and validation sets with a ratio of 8:2. All methods were adjusted to report the average best performance over three trials. The experiments were conducted on a server equipped with a GeForce 3090 GPU.
\subsection{Experimental Results}
The classification results of our method for HAR under cross-person, cross-dataset, cross-position, and one-person-to-another generalization settings are presented in Tables 1-4. We draw some conclusions from these results: 
(1) As listed in \tablename~\ref{eq:1}, in the term of average accuracy, we note that the na\"ive ERM baseline achieves favorable performance against compared arts. Most existing strategies cannot consistently improve ERM while evaluated under the rigorous settings, and some DG methods perform even worse than ERM on certain tasks, which are in well line with previous observations in \cite{gulrajani2020search}. This may be due to the inability of these methods to reduce the distribution discrepancy in time series sensor data. Therefore, it is crucial to explore domain-invariant knowledge that can effectively reduce distribution discrepancy in sensor data for HAR;
(2) Overall, our proposed approach consistently demonstrates superior performance against other state-of-the-art baselines under cross-person setting, where ours is ranked first place on all three benchmarks. Specifically, in terms of average accuracy, our CCIL significantly surpass the baseline ERM by large margins of 10.4\%, 7.3\%, and 7.4\% on DSADS, USC-HAD, and PAMAP respectively. In fact, domain generalization is a challenging task, and it is often difficult to achieve an improvement over 1\%. As can be seen in Table 1, the second-best baseline only has a slight improvement compared to the third one. In contrast to the best baseline DIVERSIFY, our approach still achieves further improvements of 2.5\%, 3.0\%, and 3.7\% on all three benchmarks. The observations validate the effectiveness of our approach compared against existing baselines;
(3) As aforementioned, other methods, such as CORAL, GroupDRO, and ANDMask, achieve competitive results on some tasks, but perform worse on others. This inconsistency may stem from their overlook for domain-invariant knowledge, potentially ignoring latent information between diverse distributions. DANN is another method for domain-invariant learning through adversarial training. It outperforms ERM in scenarios with a large number of classes, such as in the DSADS dataset. However, in cases with fewer classes, it performs even worse compared to ERM, as observed in the PAMAP2 dataset. Our method demonstrates robust performance regardless of the number of categories;
(4) As shown in Tables 2-4, it can be seen that our CCIL still consistently achieves the matched or better performance under cross-position, cross-dataset, and one-person-to-another settings. For instance, it is well known that cross-position is more challenging than other two cases. In this case, CCIL beat the best baseline by 1.3\% while achieving an improvement with about 8.1\% compared to ERM under the cross-position setting. The results demonstrate our approach has a good generalization ability for time series classification under various domain generalization evaluation settings. Importantly, our approach is very simple, that require only adding a few lines of code upon the na\"ive ERM baseline. Moreover, it is model-agonistic and can be easily integrated with other network structures for cross-domain activity recognition. Detailed results are provided in supplementary materials.
\begin{table}[t]
	\small
	\centering
	\begin{tabular}{l|cccc}
		\Xhline{1pt}
		\multirow{2}{*}{Method} & \multicolumn{4}{c}{Target} \\ \cline{2-5} 
		& \multicolumn{1}{c|}{0} & \multicolumn{1}{c|}{1} & \multicolumn{1}{c|}{2} & AVG \\ \hline
		ERM & \multicolumn{1}{c|}{51.3} & \multicolumn{1}{c|}{46.2} & \multicolumn{1}{c|}{53.1} & 50.2 \\
		Mixup & \multicolumn{1}{c|}{62.7} & \multicolumn{1}{c|}{46.3} & \multicolumn{1}{c|}{58.6} & 55.8 \\
		GroupDRO & \multicolumn{1}{c|}{51.3} & \multicolumn{1}{c|}{48.0} & \multicolumn{1}{c|}{53.1} & 50.8 \\
		RSC & \multicolumn{1}{c|}{59.1} & \multicolumn{1}{c|}{49.0} & \multicolumn{1}{c|}{59.7} & 55.9 \\
		ANDMask & \multicolumn{1}{c|}{57.2} & \multicolumn{1}{c|}{45.9} & \multicolumn{1}{c|}{54.3} & 52.5 \\
		DIVERSIFY & \multicolumn{1}{c|}{\underline{67.6}} & \multicolumn{1}{c|}{\underline{55.0}} & \multicolumn{1}{c|}{\underline{62.5}} & \underline{61.7} \\ \hline
		Ours & \multicolumn{1}{c|}{\textbf{70.2}} & \multicolumn{1}{c|}{\textbf{57.5}} & \multicolumn{1}{c|}{\textbf{63.7}} & \textbf{63.8} \\ 
		\Xhline{1pt}
	\end{tabular}
	\caption{Accuracy on one-person-to-another generalization. We use 0,1,2 to denotes the unseen test set. 0 represents DSADS, 1 represents USC-HAD, and 2 represents PAMAP2. \textbf{Bold} means the best while \underline{underline} means the second-best.}
	\label{tab:4}
\end{table}
\begin{figure}[t]
	\centering
	\subfigure[ERM]{\includegraphics[width=0.20\textwidth]{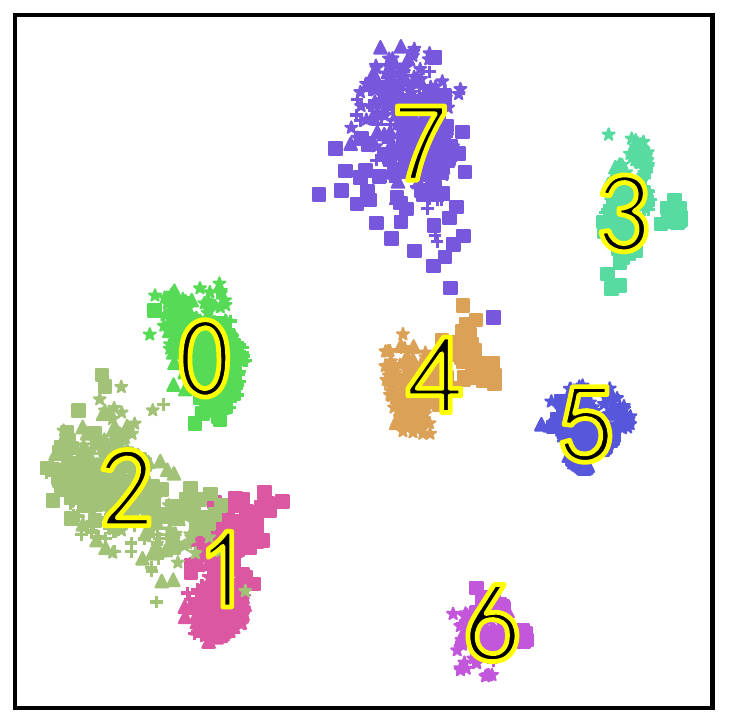}}
    \subfigure[Logit-invariance]{\includegraphics[width=0.20\textwidth]{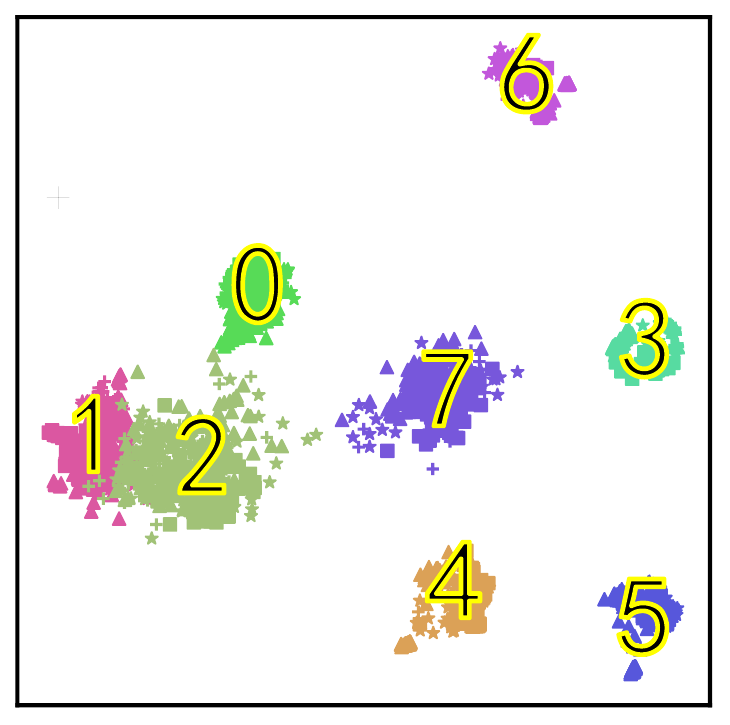}}
    \\
	\subfigure[Feature-invariance]{\includegraphics[width=0.20\textwidth]{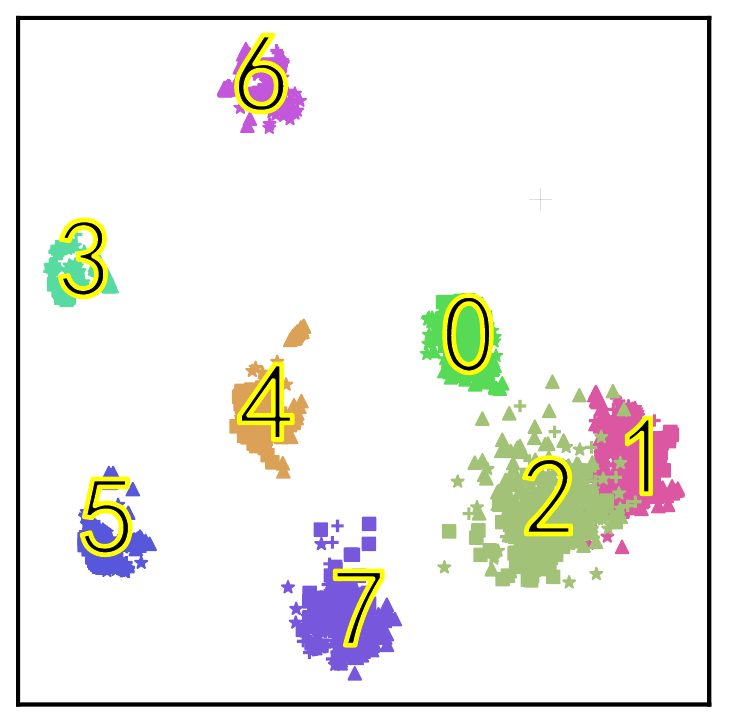}}
    \subfigure[Ours]{\includegraphics[width=0.20\textwidth]{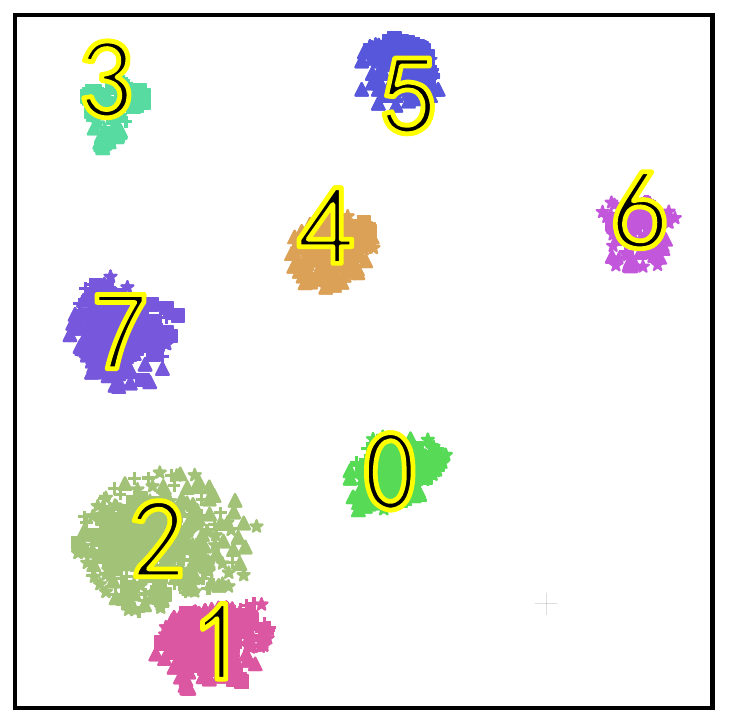}}
	\caption{Visualization of t-SNE embedding for the DSADS dataset. Here, different colors represent different classes. Different shapes indicate different domains. Best viewed in color and zoom in.}
	\label{fig2}
\end{figure}
\subsection{Ablation Study}
\label{sec:rAblation Study}
In addition to the baseline ERM method, we compare our suggested approach with the following variants to access their independent impact of each component: (1) The feature-invariance constraint abbreviated as ‘W/fea’, where Eq. \ref{eq:4} is replaced as: $\mathcal{L}_\mathrm{CMS}=\frac{1}{N_{b}}\sum_{c}\sum_{\{i|y_{i}=c\}}\|\mathbf{z}_{i}-\hat{\mathbf{z}}_{c}\|^{2}$; (2) The logit-invariance constraint abbreviated as ‘W/log’, where Eq. \ref{eq:4} is replaced as: $\mathcal{L}_\mathrm{CMS}=\frac{1}{N_{b}}\sum_{c}\sum_{\{i|y_{i}=c\}}\|\mathbf{o}_{i}-\hat{\mathbf{o}}_{c}\|^{2}$; (3) Ours with $\lambda=0$ (i.e., ‘W/$\lambda=0$’), where $\mathbf{\hat{M}}$ equals the mean value dynamical calculated from current batch; (4) Ours with $\lambda=1$ (i.e., ‘W/$\lambda=1$’), where $\mathbf{\hat{M}}$ is kept fixed from the initial pretrained model. We observe that either feature-invariance constraint or logit-invariance constraint can substantially beat the baseline ERM method. In contrast to them, our approach works the best, indicating the effectiveness of the concept matrix invariance constraint. Though the setting of $\lambda=1$ is inferior to our optimal design, it still significantly outperforms all other variants, suggesting the necessity of dynamic momentum update. 
\begin{table}[t]
    \small
	\centering
		\begin{tabular}{l|ccc|cccc|c}
			\Xhline{1pt}
			\multirow{2}{*}{Model} & \multicolumn{3}{c|}{Invariance} &\multicolumn{4}{c|}{Target (DSADS)} & \multirow{2}{*}{AVG} \\ \cline{2-8}
			& \multicolumn{1}{c|}{F}& \multicolumn{1}{c|}{L}& \multicolumn{1}{c|}{C}& \multicolumn{1}{c|}{0} & \multicolumn{1}{c|}{1} & \multicolumn{1}{c|}{2} & 3 &  \\ \hline
            ERM &\multicolumn{1}{c|}{\ding{55}} &\multicolumn{1}{c|}{\ding{55}}&\multicolumn{1}{c|}{\ding{55}}&\multicolumn{1}{c|}{83.1} & \multicolumn{1}{c|}{79.3} & \multicolumn{1}{c|}{87.8} & 71.0 & 80.3 \\ 
            \hline
            W/Fea &\multicolumn{1}{c|}{\ding{51}}&\multicolumn{1}{c|}{\ding{55}}&\multicolumn{1}{c|}{\ding{55}}& \multicolumn{1}{c|}{\underline{92.3}} & \multicolumn{1}{c|}{87.3} & \multicolumn{1}{c|}{89.1} & 85.5 & 88.6 \\
            W/Log &\multicolumn{1}{c|}{\ding{55}}&\multicolumn{1}{c|}{\ding{51}}&\multicolumn{1}{c|}{\ding{55}} &\multicolumn{1}{c|}{86.5} & \multicolumn{1}{c|}{83.6} & \multicolumn{1}{c|}{88.2} & 79.7 & 84.5 \\
            \hline
			W/$\lambda$=0 &\multicolumn{1}{c|}{\ding{55}}&\multicolumn{1}{c|}{\ding{55}}&\multicolumn{1}{c|}{\ding{51}}& \multicolumn{1}{c|}{89.1} & \multicolumn{1}{c|}{86.5} & \multicolumn{1}{c|}{88.4} & 78.0 & 85.5 \\
            W/$\lambda$=1 &\multicolumn{1}{c|}{\ding{55}}&\multicolumn{1}{c|}{\ding{55}}&\multicolumn{1}{c|}{\ding{51}}& \multicolumn{1}{c|}{92.0} & \multicolumn{1}{c|}{\underline{87.9}} & \multicolumn{1}{c|}{\underline{90.5}} & \underline{86.9} & \underline{89.3} \\
            \hline
			Ours &\multicolumn{1}{c|}{\ding{55}}&\multicolumn{1}{c|}{\ding{55}}&\multicolumn{1}{c|}{\ding{51}}& \multicolumn{1}{c|}{\textbf{94.7}} & \multicolumn{1}{c|}{\textbf{88.2}} & \multicolumn{1}{c|}{\textbf{92.5}} & \textbf{87.5} &\textbf{90.7}\\
			\Xhline{1pt}
		\end{tabular}
	\caption{Main ablation study on DSADS dataset, where ‘F’, ‘L’, and ‘C’ respectively indicates the feature-invariance, logits-invariance, and our concept matrix invariance constraints, while $\lambda$ denotes the momentum value.}
	\label{tab:5}
\end{table}
\begin{figure}[t]
	\centering
	\subfigure[DSADS]{\includegraphics[width=0.20\textwidth]{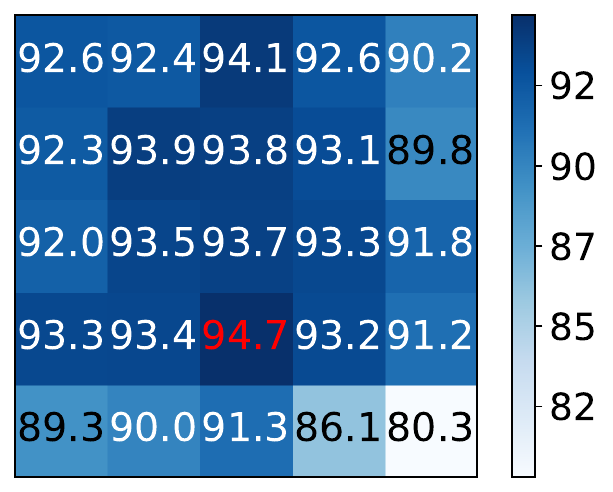}}
	\subfigure[USC-HAD]{\includegraphics[width=0.20\textwidth]{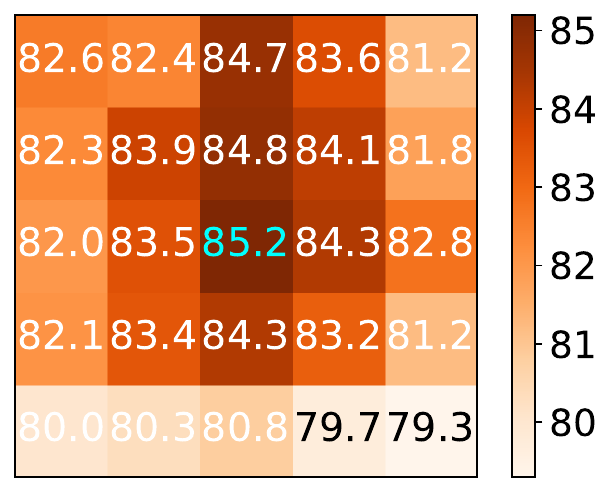}}
	\caption{Parameters sensitivity analysis of ${\alpha}$ and ${\lambda}$. The horizontal axis signifies $\alpha \in \{0.1, 0.5, 1, 5, 10\}$, while the vertical axis denotes $\lambda \in \{0, 0.9, 0.99, 0.999, 0.9999\}$.}
	\label{fig3}
\end{figure}
\subsection{T-SNE Visualization}
To better understanding our invariance regularization, we provide a t-SNE visualization illustration on DSADS dataset, as plotted in \figurename~\ref{fig2}. In contrast to the other three strategies, it can be seen the clusters from our proposed invariance regularization are more distinctly separated, indicating its effectiveness while generalizing on an unseen distribution. This is in well line with the results reported in Table 5. Meanwhile, we observe that the logit-invariance constraint alone performs only slightly better than the baseline ERM method, both of which are inferior to the feature-invariance constraint. This is not surprising that since the logit only can provide a coarse value, which is incapable of capturing fine-grained domain-invariant representations. Therefore, the feature-invariance constraint can provide a more subtle representation compared to both of them. However, due to ignoring the effect of classifier weights, the feature-invariance constraint possibly causes the model to concentrate on unimportant features. In contrast, our CCIL can produce a more robust and stable clustering results.
\subsection{Parameter Sensitivity Analysis}
We focus on the momentum coefficient $\lambda$ and the parameter $\alpha$ in CCIL, which are empirically evaluated for their sensitivities by choosing values from $\{0, 0.9, 0.99, 0.999, 0.9999\}$ and $\{0.1, 0.5, 1, 5, 10\}$, respectively. The results are shown in \figurename~\ref{fig3}. It can be seen that the CCIL method has robust performance across a wide range of hyperparameters on the DSADS and USC-HAD datasets. From the results, we observed that the performance is inferior when $\lambda = 0$ compared to values such as $\lambda = 0.9$. The best performance is achieved when $\lambda = 0.9$ and $\alpha = 1$. This indicates that they play a crucial role in generalization performance, necessitating the use of a momentum update strategy for updating the concept matrix.
\section{Conclusion}
In this paper, we propose CCIL, a new regularization approach for sensor-based cross-domain activity recognition. While there exist diverse distributions in time series activity data across domains, e.g., different persons, CCIL addresses this problem by learning domain-invariant knowledge. To ensure robust outputs, the key idea of our algorithm is to capture domain-invariant knowledge by enforcing similarity between the concept matrix of samples from the same activity category and their corresponding mean value. Different from prior most works, our approach takes a different path by taking into full consideration the classifier weights (i.e., the logit-invariance), rather than only concentrating on feature-invariance. Experiments on multiple public datasets demonstrate the superiority of our CCIL approach across various cross-domain settings.

\section{Supplementary Materials}
In the Supplementary Material, all experiments were conducted under the CNN architecture (except for the extensibility section) and we provide: 
\begin{itemize}
	\item Different invariance regularizations.
	\item Implementation details.
    \item Extensibility.
	\item Additional visualization study.
	\item Pseudo-code.
\end{itemize}

\appendix
\section{Different Invariance Regularizations}
\begin{figure}[t]
	\centering
	\subfigure[Logit-invariance]{\includegraphics[width=0.185\textwidth]{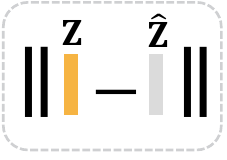}}
	\subfigure[Feature-invariance]{\includegraphics[width=0.255\textwidth]{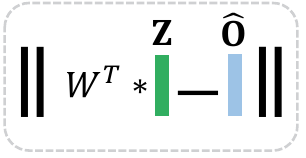}}
	\\
	\subfigure[Ours]{\includegraphics[width=0.44\textwidth]{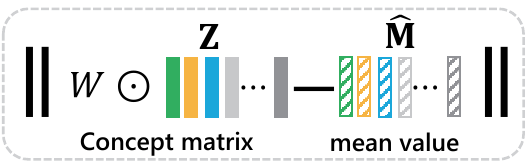}}
	\caption{Visualized versions of different invariance regularization.}
	\label{fig5}
\end{figure}

\figurename~\ref{fig5} presents visualized versions of different invariance regularization as mentioned in the ablation study of the main paper. Here $\Vert \cdot \Vert$ are the $l_2$ norm; Subfigures (a) - (c) denote the Logit-invariance, Feature-invariance and CCIL; $\mathbf{W}$ is the weight in the classifier; $\odot$ is the element-wise product.; $\mathbf{z}$ is the feature; $\mathbf{o}$ is the logit; $\hat{\mathbf{M}}$ is the mean value; $\mathbf{W \odot \mathbf{z}}$ is the concept matrix. Unlike feature and logit invariance regularization, our Categorical Concept Invariant Learning considers both feature and classifier weights, enabling effective learning of domain-invariant information.
\section{Implementation Details}

\subsection{Datasets}
The following provides a detailed description of four publicly available HAR benchmark datasets. The UCI Daily and Sports Dataset (DSADS) \cite{altun2010comparative} consists of 19 activities collected from 8 subjects wearing body-worn sensors on 5 body parts. The subjects were aged between 20 and 30 years. The USC-SIPI Human Activity Dataset (USC-HAD) \cite{zhang2012usc} is composed of 14 subjects (7 male, 7 female, aged 21 to 49) executing 12 activities with a sensor tied to the front right hip. The UCI-HAR \cite{anguita2013public} dataset was collected from 30 subjects performing 6 daily living activities with a waist-mounted smartphone. The subjects were aged between 20 and 30 years. The PAMAP dataset \cite{reiss2012introducing} contains data on 18 activities, performed by 9 subjects wearing 3 sensors. The subjects were aged between 20 and 30 years. Additional information is provided in \tablename~\ref{tab:my-table-dataset}.

\begin{table}[htbp]
	\small
    \setlength{\tabcolsep}{3pt}
	\centering
	\begin{tabular}{l|cccc}
		\Xhline{1pt}
		Dataset & \multicolumn{1}{c|}{Subject} & \multicolumn{1}{c|}{Age } & \multicolumn{1}{c|}{Sampling} & \multicolumn{1}{c}{Sampling rate}  \\\hline
		DSADS & \multicolumn{1}{c|}{8} & \multicolumn{1}{c|}{20-30} & \multicolumn{1}{c|}{1,140,000} & \multicolumn{1}{c}{25 Hz}  \\ 
		PAMAP2 & \multicolumn{1}{c|}{9} & \multicolumn{1}{c|}{23-31} & \multicolumn{1}{c|}{3,850,505} & \multicolumn{1}{c}{100 Hz}\\ 
		USC-HAD & \multicolumn{1}{c|}{14} & \multicolumn{1}{c|}{21-49} & \multicolumn{1}{c|}{5,441,000} & \multicolumn{1}{c}{100 Hz}  \\ 
		UCI-HAR & \multicolumn{1}{c|}{30} & \multicolumn{1}{c|}{19-48} & \multicolumn{1}{c|}{ 1,310,000} & \multicolumn{1}{c}{50 Hz}  \\ 
		\Xhline{1pt}
	\end{tabular}
	\caption{Statistical information of four HAR datasets.}
	\label{tab:my-table-dataset}
\end{table}

\begin{table}[htbp]
	\small  
	\centering
	\begin{tabular}{l|ccc}
		\Xhline{1pt}
		Setting & \multicolumn{1}{c|}{Dataset} & \multicolumn{1}{c|}{Input } & \multicolumn{1}{c}{Kernel Size}   \\\hline
		\multirow{3}{*}{Cross-person} & \multicolumn{1}{c|}{DSADS} & \multicolumn{1}{c|}{(45,1,125)} & \multicolumn{1}{c}{(1,9)}  \\ 
		& \multicolumn{1}{c|}{PAMAP2} & \multicolumn{1}{c|}{(27,1,200)} & \multicolumn{1}{c}{(1,9)} \\ 
		& \multicolumn{1}{c|}{USC-HAD} & \multicolumn{1}{c|}{(6,1,200)} & \multicolumn{1}{c}{(1,6)}  \\ \hline 
		Cross-position & \multicolumn{1}{c|}{DSADS} & \multicolumn{1}{c|}{(9,1,125)} & \multicolumn{1}{c}{(1,9)}   \\  \hline
		Cross-dataset & \multicolumn{1}{c|}{-} & \multicolumn{1}{c|}{(6,1,50)} & \multicolumn{1}{c}{(1,6)}   \\ 
		\Xhline{1pt}
	\end{tabular}
	\caption{Information on the architectures of the models.}
	\label{tab:my-table-kernelsize}
\end{table}

\subsection{Network Architecture and Training Details}
The model architecture consists of two blocks, each comprising a convolution layer, a pooling layer, and a batch normalization layer. A single fully connected layer serves as the classifier. All methods are implemented using PyTorch. The maximum number of training epochs is set to 150. The Adam optimizer with a weight decay of $5 \times 10^{-4}$ is employed. The learning rate for GILE is $10^{-4}$, while for other methods, it is either $10^{-2}$ or $10^{-3}$. We tune the hyperparameters for each method individually. For the pooling layer, MaxPool2d from PyTorch is used, with a kernel size of (1, 2) and a stride of 2. For the convolution layer, Conv2d from PyTorch is employed. Different tasks use different kernel sizes, as shown in \tablename~\ref{tab:my-table-kernelsize}.

\begin{table}[t]
	\small  
	\centering
	\begin{tabular}{l|ccccc}
		\Xhline{1pt}
		\multirow{2}{*}{Method} & \multicolumn{5}{c}{Target (DSADS)} \\ \cline{2-6} 
		& \multicolumn{1}{c|}{0} & \multicolumn{1}{c|}{1} & \multicolumn{1}{c|}{2} & \multicolumn{1}{c|}{3} & AVG \\ \hline
		ERM & \multicolumn{1}{c|}{88.0} & \multicolumn{1}{c|}{84.2} & \multicolumn{1}{c|}{88.0} & \multicolumn{1}{c|}{75.3} & 83.4 \\
		DANN & \multicolumn{1}{c|}{89.4} & \multicolumn{1}{c|}{85.3} & \multicolumn{1}{c|}{86.1} & \multicolumn{1}{c|}{83.9} &  86.2 \\
		CORAL & \multicolumn{1}{c|}{91.2} & \multicolumn{1}{c|}{85.2} & \multicolumn{1}{c|}{86.8} & \multicolumn{1}{c|}{79.3} &  85.6 \\
		Mixup & \multicolumn{1}{c|}{89.8} & \multicolumn{1}{c|}{83.4} & \multicolumn{1}{c|}{90.1} & \multicolumn{1}{c|}{\underline{87.1}} & 87.6 \\
		GroupDRO & \multicolumn{1}{c|}{92.1} & \multicolumn{1}{c|}{86.7} & \multicolumn{1}{c|}{89.3} & \multicolumn{1}{c|}{78.4} &  86.6 \\
		RSC & \multicolumn{1}{c|}{85.6} & \multicolumn{1}{c|}{82.4} & \multicolumn{1}{c|}{87.9} & \multicolumn{1}{c|}{78.0} & 83.4 \\
		ANDMask & \multicolumn{1}{c|}{85.5} & \multicolumn{1}{c|}{76.1} & \multicolumn{1}{c|}{88.2} & \multicolumn{1}{c|}{78.2} & 82.2 \\
		DIVERSIFY & \multicolumn{1}{c|}{\underline{91.3}} & \multicolumn{1}{c|}{\underline{86.9}} & \multicolumn{1}{c|}{\underline{90.7}} & \multicolumn{1}{c|}{86.3} &  \underline{88.8} \\ \hline
		Ours & \multicolumn{1}{c|}{\textbf{95.2}} & \multicolumn{1}{c|}{\textbf{92.7}} & \multicolumn{1}{c|}{\textbf{93.6}} & \multicolumn{1}{c|}{\textbf{89.3}} &\textbf{92.7} \\ 
		\Xhline{1pt}
	\end{tabular}
	\caption{Accuracy on cross-person generalization. We use 0,1,2,3 to denotes the unseen test set. \textbf{Bold} means the best while \underline{underline} means the second-best.}
	\label{tab:8}
\end{table}
\begin{figure}[t]
	\centering
	\subfigure[ERM]{\includegraphics[width=0.20\textwidth]{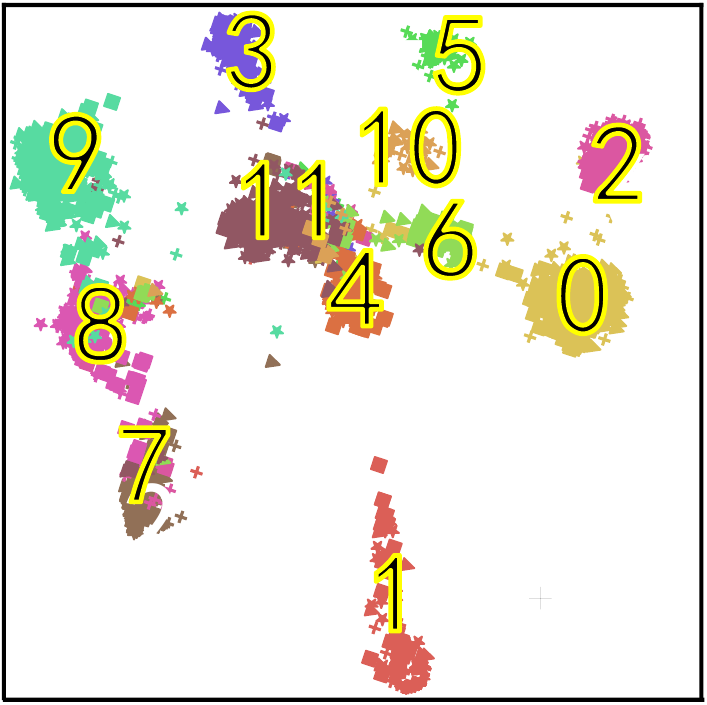}}
	\subfigure[Logit-invariance]{\includegraphics[width=0.20\textwidth]{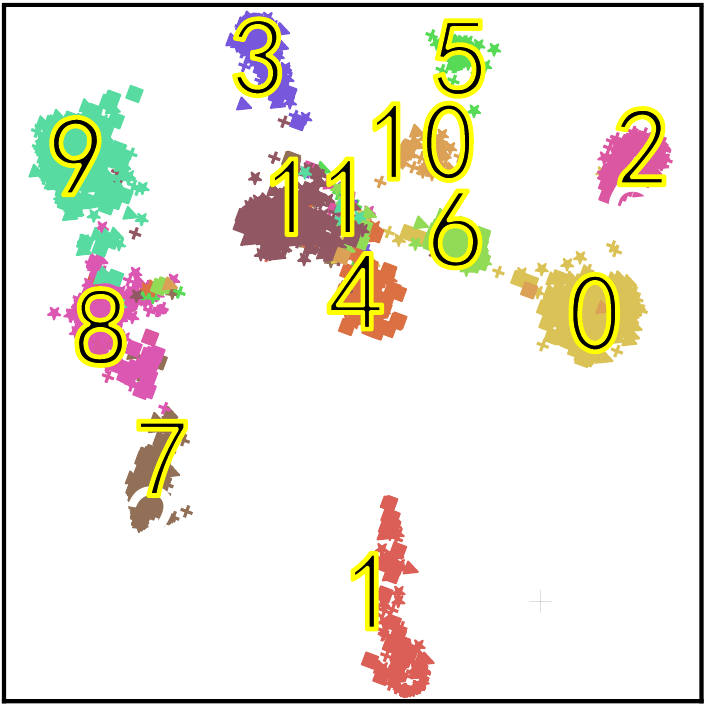}}
	\\
	\subfigure[Feature-invariance]{\includegraphics[width=0.20\textwidth]{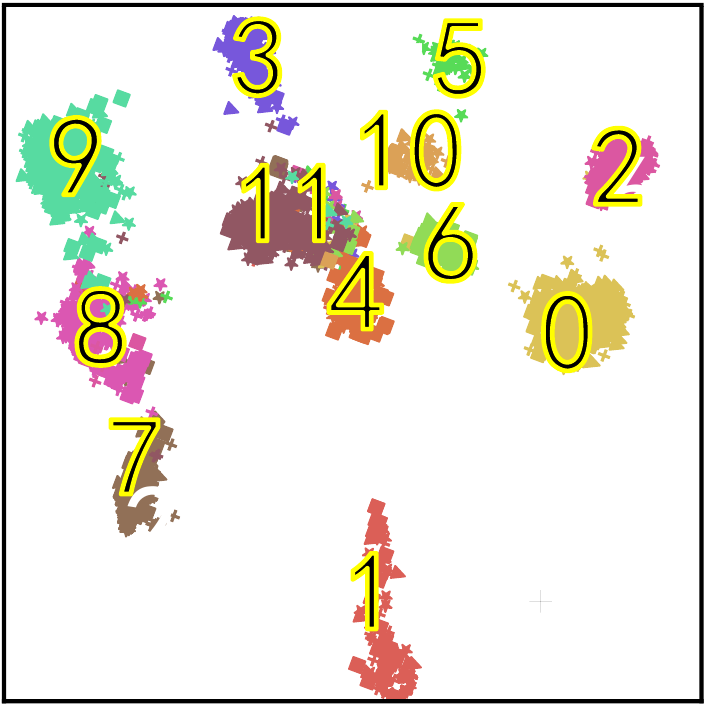}}
	\subfigure[Ours]{\includegraphics[width=0.20\textwidth]{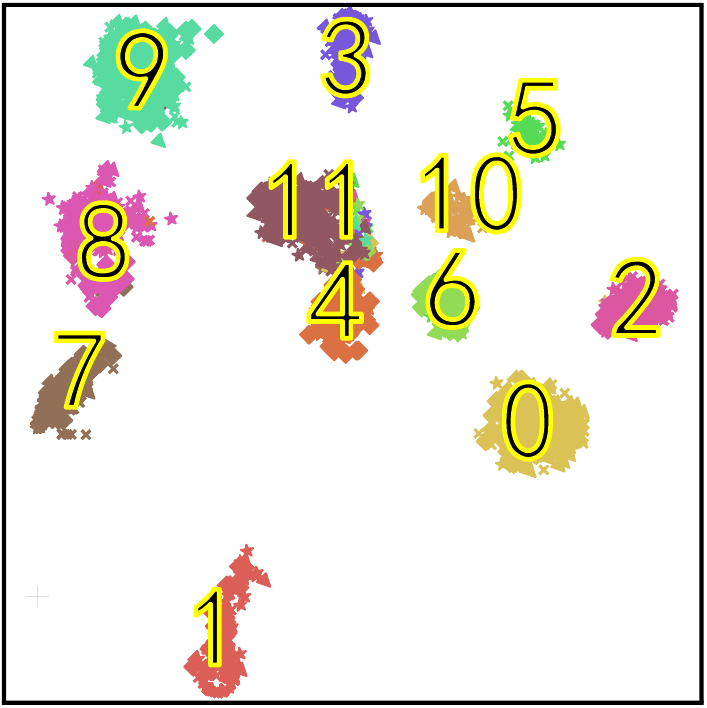}}
	\caption{Visualization of t-SNE embedding for the USC-HAD dataset. Here, different colors represent different classes. Different shapes indicate different domains. Best viewed in color and zoom in.}
	\label{fig6}
\end{figure}

\begin{algorithm}[!t]
	\caption{CCIL for DG-based HAR}
	\renewcommand{\algorithmicrequire}{\textbf{Input:}} 
	\renewcommand{\algorithmicensure}{\textbf{Output:}}
	\raggedright
	\underline{\textbf{\emph{Training:}}}
	\begin{algorithmic}[1]
		\REQUIRE
		The training domain $\mathcal{D}^{tr}$, hyperparameter $\alpha$ and the momentum  $\lambda$
		\ENSURE
		The parameters $\mathbf{\theta}$ of feature extractor $f$ and the weights $\mathbf{W}$ of activity classifier $g$.\\
		\STATE Randomly initialize $\mathbf{\theta}$ and $\mathbf{W}$;
		\WHILE{not converge}{
			\STATE Sample a mini-batch $\mathcal{B} \gets\{\mathcal{B}^{1},\mathcal{B}^{2},...,\mathcal{B}^{n} \}$ from each source domain, and concat them as $x_{tr}$;
			\STATE Extract output features $\mathbf{z} \gets f (\mathbf{x}_{tr})$ by feature extractor $f$;
			\STATE Obtain the classification loss $\mathcal{L}_\mathbf{CE}$ for activity classifier $g$;
			\STATE Using output features $\mathbf{z}$ and classifier weight $\mathbf{W}$ to calculate the mean values $\mathbf{\hat{M}}$;
			\STATE Updating the Concept matrix $\mathbf{M}$ through dynamic Momentum Update Strategy with momentum value $\lambda$, i.e., Eq. (5).\\
			\STATE Calculate the loss of CMS ($\mathcal{L}_\mathbf{CMS}\gets\frac{1}{N_{b}}\sum_{c}\sum_{\{i|y_{i}=c\}}\|\mathbf{M}_{i}-\hat{\mathbf{M}}_{c}\|^{2}$);\\
			\STATE Calculate the final loss of CCIL ($\mathcal{L} \gets \mathcal{L}_\mathbf{CE} + \alpha \mathcal{L}_\mathbf{CMS}$) ;
			\STATE Update $\mathbf {\theta}$ and $\mathbf{W}$ using Adam;
		}
		\ENDWHILE\\
	\end{algorithmic}
	\raggedright
	\underline{\textbf{\emph{Inference:}}}
	\begin{algorithmic}[1]
		\REQUIRE
		The trained feature extractor $f$ and classifier $g$, test domain data $\mathcal{D}^{te}$.\\
		\ENSURE
		Classification results on the test domain.\\
		\FOR{$(x,y) \in \mathcal{D}^{te}$}
		\STATE Get the predict label $\overline{y} \gets g(f(x))$;
		\ENDFOR
		\STATE Calculate the classification accuracy.
		\RETURN Classification results on target sensor data.
	\end{algorithmic}
\end{algorithm}	
\subsection{Compared Methods}
The following provides a detailed description of various comparison methods. \textbf{ERM} \cite{vapnik1991principles} is a popular DG-based method that focuses on minimizing the sum of errors over source domains. \textbf{DANN} \cite{ganin2016domain} is a method that utilizes the adversarial training to force the discriminator unable to classify domains for better domain-invariant features. It requires domain labels and splits data in advance while ours is a universal method. \textbf{CORAL} \cite{sun2016deep} is a method that utilizes the covariance alignment in feature layers for better domain-invariant features. It also requires domain labels and splits data in advance. \textbf{Mixup} \cite{zhang2018mixup} is a method that utilizes interpolation to generate more data for better generalization. Ours mainly focuses on generalized representation learning. \textbf{GroupDRO} \cite{sagawadistributionally} is a method that seeks a global distribution with the worst performance within a range of the raw distribution for better generalization. Ours study the internal distribution shift instead of seeking a global distribution close to the original one. \textbf{RSC} \cite{huang2020self} is a self-challenging training algorithm that forces the network to activate features as much as possible by manipulating gradients. It belongs to gradient operation-based DG while ours is to learn generalized features. \textbf{ANDMask} \cite{parascandololearning} is another gradient-based optimization method that belongs to special learning strategies. Ours focuses on representation learning. \textbf{GILE} \cite{qian2021latent} is a disentanglement method designed for cross-person human activity recognition. It is based on VAEs and requires domain labels. \textbf{AdaRNN} \cite{du2021adarnn} is a method with a two-stage that is non-differential and it is tailored for RNN. A specific algorithm is designed for splitting. Ours is universal and is differential with better performance. \textbf{DIVERSIFY} \cite{lu2024diversify} is an adversarial training method designed to maximize the potential distributional scenarios of the "worst case" and then minimize the resulting distributional differences.
\section{Extensibility}
We explore using Transformer as the backbone for comparison. Transformers often exhibit better generalization ability compared to CNNs, making further improvement with Transformers more challenging. As shown in \tablename~\ref{tab:8}, each method with a Transformer backbone shows significant improvement on DSADS. While DANN shows little improvement over ERM, our method still achieves further enhancements and delivers the best performance. Overall, across all architectures, our method consistently achieves superior performance.
\section{Additional Visualization Study}
We provide additional visualization analysis in this section. As illustrated in \figurename~\ref{fig6}, we performed T-SNE embedding on USC-HAD, and it is clear that neither ERM, Logit-invariance, nor Feature-invariance methods achieve satisfactory domain-invariant representations. Although Logit-invariance and Feature-invariance methods perform better than ERM, their effectiveness remains limited. In contrast, our proposed method consistently achieves superior domain-invariant representations.

\section{Pseudo-Code}
To better understand the CCIL algorithm, we have provided pseudo-code for the training and inference processes of the CCIL algorithm in this paper.

\section{Contribution Statement}
\begin{itemize}
	\item \textbf{Di~Xiong~(Nanjing Normal University, China):} Proposing the idea, implementing code, conducting experiments, data collection, figure drawing, table organizing, and completing original manuscript.
	\item \textbf{Shuoyuan~Wang~(Southern University of Science and Technology, China):}  Writing polish, idea improvement, and rebuttal assistance.
	\item \textbf{Lei~Zhang~(Nanjing Normal University, China):} Providing experimental platform, supervision, idea improvement, writing polish, rebuttal assistance, and funding acquisition.
    \item \textbf{Wenbo~Huang~(Southeast University, China):} Figure drawing, table organizing, supervision, and rebuttal assistance.
    \item \textbf{Chaolei~Han~(Southeast University, China):} Table organizing.
\end{itemize}

\section{Acknowledgements}
The authors would like to appreciate all participants of peer review. The work was supported in part by the National Natural Science Foundation of China under Grant 62373194, in part by the Excellent Ph.D Training Program (SEU).

\bibliography{aaai25}

\end{document}